%% file: main.tex
\pdfmapfile{+ps2pk35.map}
\pdfmapfile{+cm.map}
\pdfmapfile{+symbols.map}
\documentclass[11pt,onecolumn,letterpaper]{article}
\AddToHookNext{shipout/foreground}{%
  \begin{tikzpicture}[remember picture,overlay]
    \node[anchor=north east,inner sep=0pt]
      at ([xshift=-1in,yshift=-0.30in]current page.north east) {%
        \raisebox{-0.5\height}{\includegraphics[width=0.82in,height=0.29in,keepaspectratio]{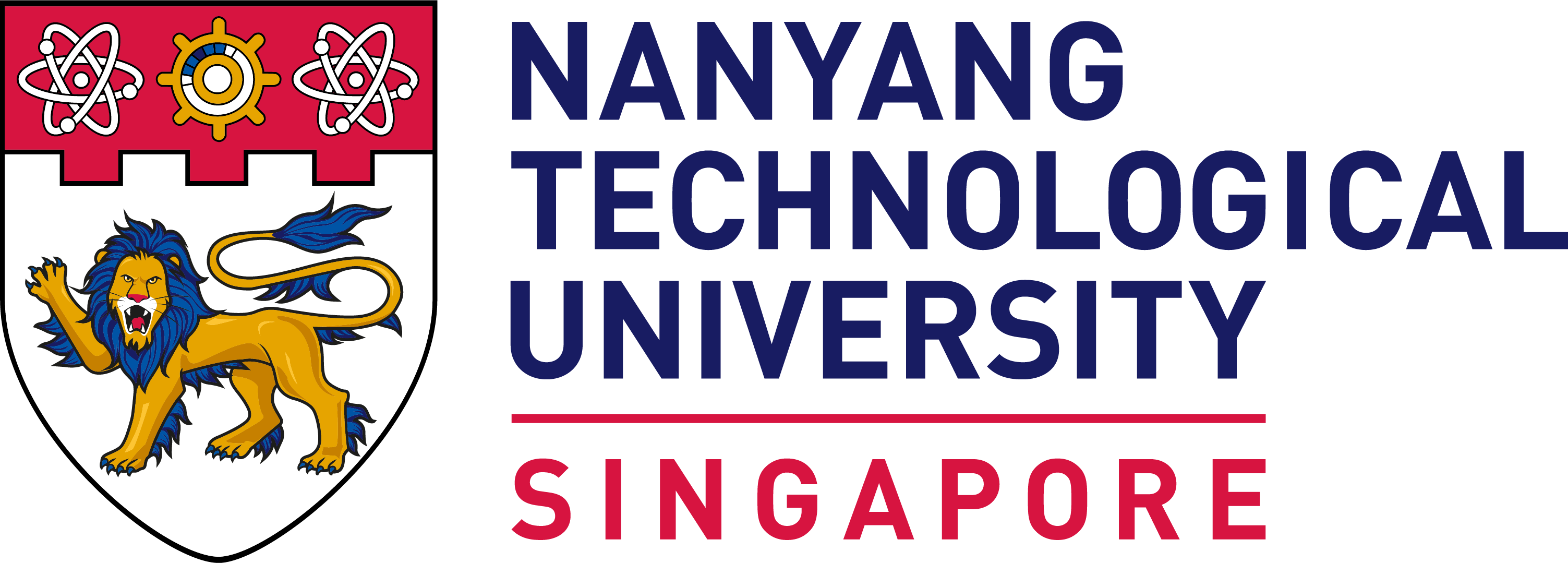}}%
        \hspace{0.14in}%
        \begingroup
          \sbox{0}{\includegraphics{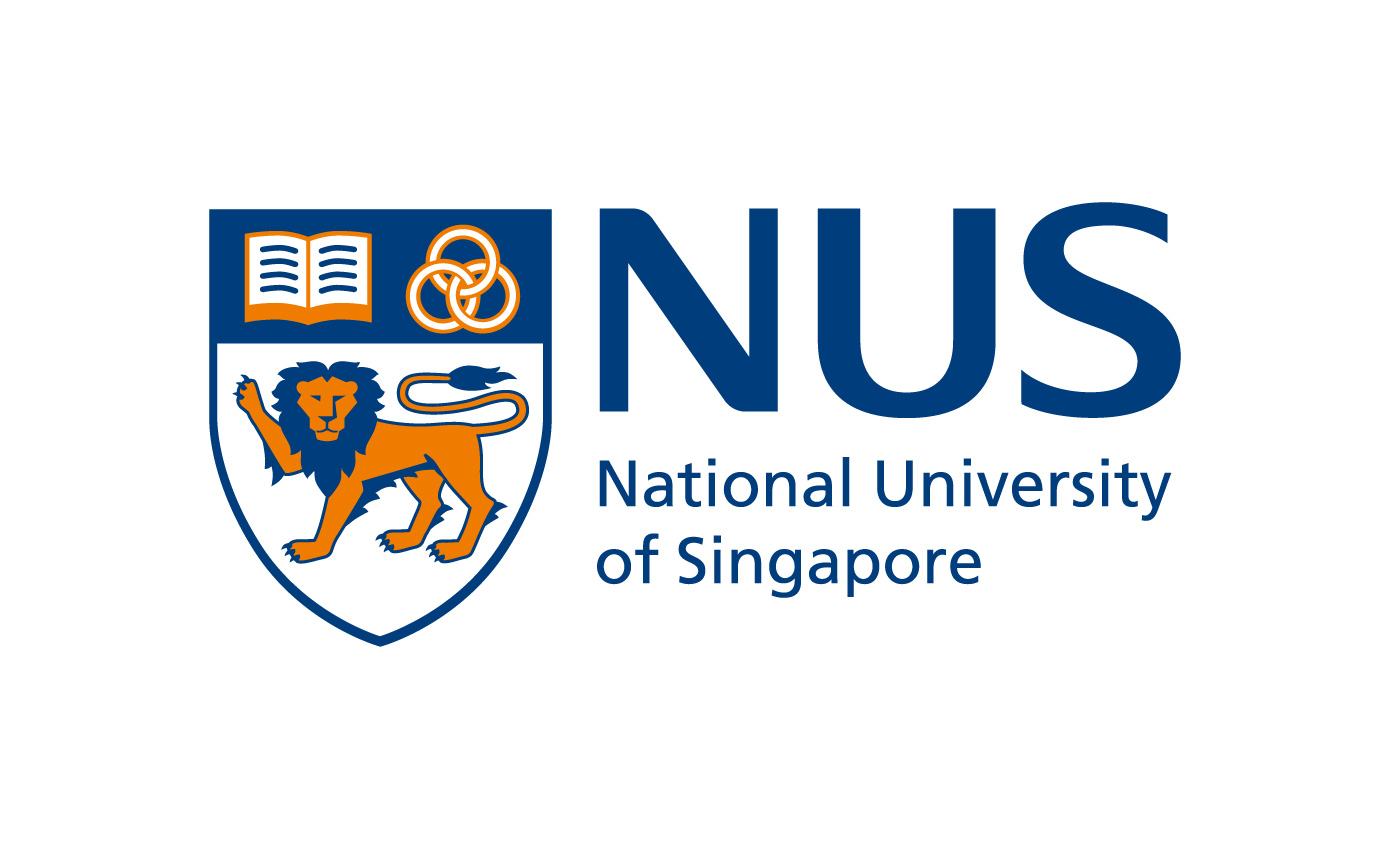}}%
          \raisebox{-0.5\height}{\includegraphics[
            trim=.145\wd0 .24\ht0 .145\wd0 .23\ht0,clip,
            width=0.82in,height=0.31in,keepaspectratio]{nus-logo.jpg}}%
        \endgroup%
        \hspace{0.22in}%
        \raisebox{-0.5\height}{\includegraphics[width=0.96in,height=0.30in,keepaspectratio]{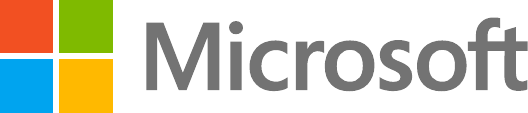}}%
      };
    \draw[line width=0.6pt,color=black!70] ([xshift=1in,yshift=-0.70in]current page.north west) -- ([xshift=-1in,yshift=-0.70in]current page.north east);
  \end{tikzpicture}%
}

\usepackage[pagenumbers]{cvpr}
\usepackage{graphicx,amsmath,amssymb,booktabs,array,microtype}
\usepackage[tighter]{newtxtext}
\usepackage{newtxmath}
\usepackage{tabularx,array,amssymb,microtype}
\usepackage{colortbl}
\definecolor{rwheader}{RGB}{242,237,248}
\definecolor{rwours}{RGB}{253,239,224}
\newcommand{\cmark}{\ensuremath{\checkmark}}
\newcommand{\xmark}{\ensuremath{\times}}
\usepackage{balance,placeins}
\usepackage[accsupp]{axessibility}
\usepackage{xcolor}

\usepackage{tikz}
\usetikzlibrary{arrows.meta}
\usepackage{pgfplots}
\pgfplotsset{compat=1.17}
\usepackage{tcolorbox}
\tcbuselibrary{breakable}
\definecolor{seekdark}{HTML}{3A3133}
\definecolor{seekpolicy}{HTML}{6A4A7C}
\definecolor{seekbank}{HTML}{B9623C}
\DeclareRobustCommand{\seeklogo}{%
  \begin{tikzpicture}[baseline=(S.base),
    letter/.style={anchor=base west,inner sep=0pt,outer sep=0pt,font={\sffamily\bfseries}},
    loop/.style={line width=0.06em,line cap=round,overlay,-{Stealth[length=0.24em,width=0.22em]}}]
    \node[letter,text=seekdark] (S) at (0,0) {S};
    \node[letter,text=seekpolicy] (E1) at (S.base east) {E};
    \node[letter,text=seekbank] (E2) at (E1.base east) {E};
    \node[letter,text=seekdark] (K) at (E2.base east) {K};
    \draw[loop,seekpolicy] ([yshift=0.10em]E1.north) to[bend left=55] ([yshift=0.10em]E2.north);
    \draw[loop,seekbank] ([yshift=-0.10em]E2.south) to[bend left=55] ([yshift=-0.10em]E1.south);
  \end{tikzpicture}%
}

\definecolor{linkblue}{RGB}{37,85,123}
\usepackage[breaklinks,colorlinks,allcolors=linkblue]{hyperref}
\usepackage[skip=6.5pt plus 1pt,indent=0pt]{parskip}
\newcommand{\seek}{SEEK}
\newcommand{\papertitle}{\texorpdfstring{\seeklogo}{SEEK}: Self-Evolution through Execution Knowledge\texorpdfstring{\\}{ }for Agentic Robot Manipulation}
\title{\papertitle}
\author{%
Zeyang Guan$^{1}$\quad Qihao Zhao$^{1}$\quad Rui Gong$^{1}$\quad Zhang Haitian$^{1}$\quad Yalun Dai$^{1}$\\
Yangyu Huang$^{2}$\quad Lingao Xiao$^{3}$\quad Haozhe Du$^{3}$\quad Xuehui Yu$^{3}$\quad Ye Lu$^{1}$\quad Jianjun Gao$^{1}$\\
Ruoyu Wang$^{1}$\quad Haowen Bai$^{1}$\quad Hao Li$^{1}$\quad Scarlett Li$^{2}$\quad Kim-Hui Yap$^{1}$\\[5pt]
{\small\mdseries $^{1}$Nanyang Technological University\quad $^{2}$Microsoft Research\quad $^{3}$National University of Singapore}\\[6pt]
{\small\bfseries\color{seekpolicy}\href{https://zeyangguan.github.io/SEEK-Agentic-Robot-Manipulation/}{\color{seekpolicy}Project Page}\hspace{1.2em}\href{https://github.com/ZeyangGuan/SEEK-Agentic-Robot-Manipulation}{\color{seekpolicy}GitHub}}%
}
\hypersetup{pdftitle={\papertitle}}
\hypersetup{pdfauthor={Zeyang Guan, Qihao Zhao, Rui Gong, Zhang Haitian, Yalun Dai, Yangyu Huang, Lingao Xiao, Haozhe Du, Xuehui Yu, Ye Lu, Jianjun Gao, Ruoyu Wang, Haowen Bai, Hao Li, Scarlett Li, Kim-Hui Yap}}

\makeatletter
\def\@maketitle{%
\newpage
\null
\vspace*{-2.5pt}
\begin{center}
{\fontsize{20}{25}\selectfont\bfseries \@title\par}
\vskip 19.5pt
{\fontsize{10.95}{16}\selectfont\bfseries\lineskip .5em
\begin{tabular}[t]{@{}c@{}}\@author\end{tabular}\par}
\vskip 28pt
\end{center}
}
\def\cvprsection{\@startsection{section}{1}{\z@}{-17pt plus -3pt minus -2pt}{9pt plus 1pt}{\fontsize{13}{16}\selectfont\bfseries}}
\def\cvprsubsection{\@startsection{subsection}{2}{\z@}{-14pt plus -2pt minus -1pt}{3.5pt}{\large\bfseries}}
\renewcommand\paragraph{\@startsection{paragraph}{4}{\z@}{6.5pt plus 1pt}{-1em}{\normalsize\bfseries}}
\makeatother

\begin{document}
\frenchspacing
\raggedbottom
\clubpenalty=10000
\widowpenalty=10000
\clubpenalties=2 10000 0
\displaywidowpenalty=10000
\lefthyphenmin=2
\righthyphenmin=3
\maketitle

\input{sections/00_abstract}
\suppressfloats[t]
\begin{figure}[t]
\centering
\includegraphics[width=\linewidth]{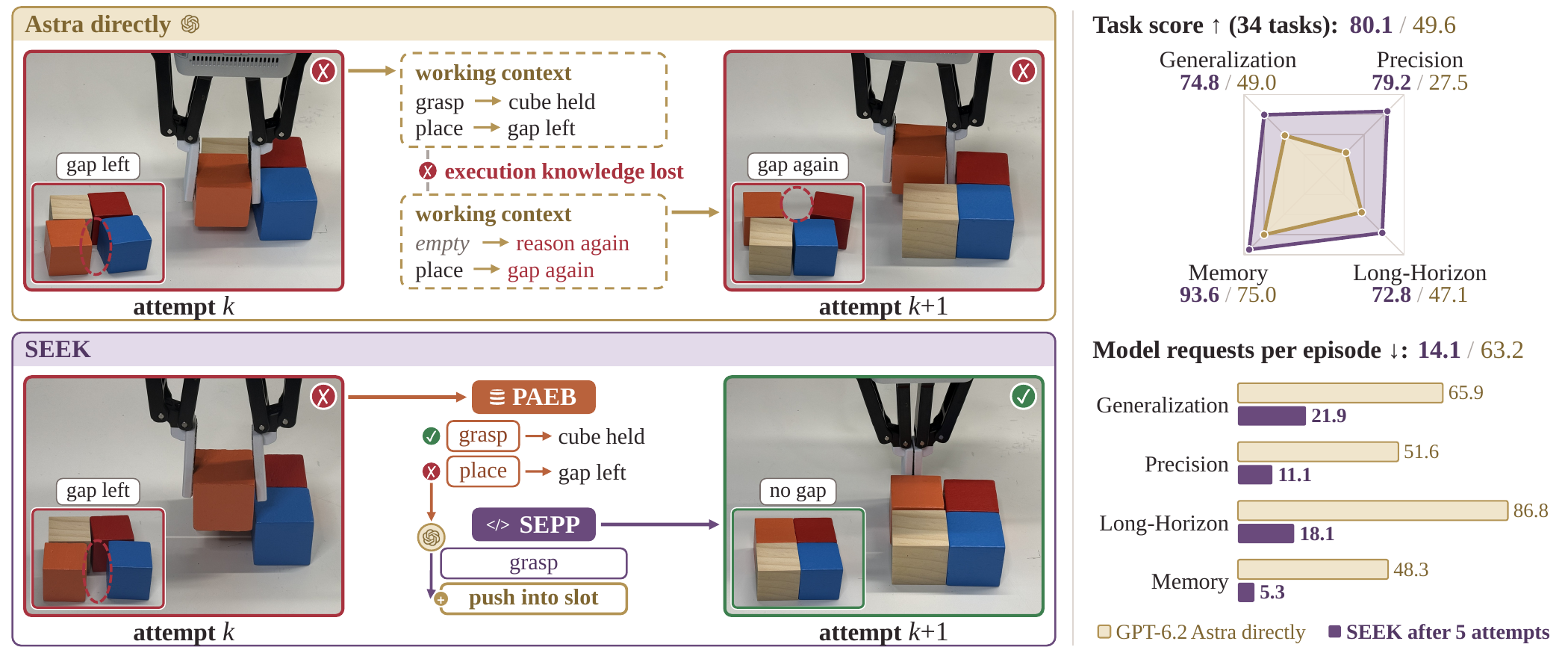}
\caption{\textbf{Motivation and central idea of \seek.} Top: GPT-6 Astra~\cite{gpt6astra2026} acting directly keeps what the robot learns only in its working context and loses it once the task ends, so the next attempt may repeat the same mistakes; on a real robot, the last cube of a square is again grasped and set down, leaving a gap. Bottom: \seek{} retains this execution knowledge across attempts as action-effect evidence (PAEB), including that of failed attempts, and reusable procedures (SEPP), while the model remains fixed; its next attempt pushes the cube in. Right: with GPT-6.2 Astra on 34 RoboDojo tasks, task score by category and model requests per episode of \seek{} after 5 attempts and of the model acting directly~\cite{roboprobe2026}. 
    }
    \label{fig:framework}
\end{figure}
\input{sections/01_introduction}
\input{sections/02_related}
\input{sections/04_method}
\input{sections/05_evidence}
\input{sections/07_conclusion}

\nobalance
{\small\bibliographystyle{ieeenat_fullname}\bibliography{references}}
\end{document}

%% file: sections/00_abstract.tex
\definecolor{seekcontribbg}{HTML}{F3ECFD}
\begin{tcolorbox}[breakable, colback=seekcontribbg, colframe=seekcontribbg, boxrule=0pt, arc=2pt,
  boxsep=0pt, left=14pt, right=14pt, top=10pt, bottom=7pt, before skip=0pt, after skip=6pt]
\centerline{\large\bfseries Abstract}
\vspace{4pt}
{\itshape
    General-purpose multimodal foundation models enable robots to adapt during deployment from observations and execution feedback, without task-specific demonstrations or training. However, the robot retains what it learns \textbf{only in the model’s working context}. Once the task ends, it loses this execution knowledge, such as how to carry a bowl between two grippers. On the next attempt at the task, the robot may take just as long and repeat the same mistakes. To avoid this repetition, robots must \textbf{retain execution knowledge across attempts}. Yet replaying past actions is insufficient, because their effects depend on the current conditions. Procedures that determine actions from the current scene address this, but when they meet conditions that earlier attempts did not cover, revising them requires the measured effects of earlier actions. We therefore propose \textbf{SEEK}, an agentic framework that retains both, enabling robots to improve across attempts while keeping the underlying model fixed. To adapt retained knowledge to the current scene, SEEK adopts a \textbf{Self-Evolving Programmatic Policy (SEPP)} that encodes manipulation as executable procedures and determines each action from current observations. As the robot executes these procedures, SEEK records the measured effects of its actions, including those from failed attempts, in a \textbf{Persistent Action-Effect Bank (PAEB)}. Where a retained procedure applies, it acts without a model request; where it meets a condition it does not cover, this evidence guides its revision. On RoboDojo, knowledge retained over ten attempts raises SEEK’s average task score from 27.0 to \textbf{54.4} and its success rate from 19.7\% to \textbf{48.0\%} across all 42 tasks. On 34 of these tasks with GPT-6.2 Astra, SEEK reaches \textbf{80.1} after five attempts, against 49.6 for the same model acting directly, and uses \textbf{77.6\%} fewer model requests.\par}
\end{tcolorbox}

%% file: sections/01_introduction.tex
\section{Introduction}
Recent advances in robot learning have been driven by scaling demonstration data and model capacity, producing policies with broader task coverage and stronger generalization~\cite{rt12023,rtx2024,rt22023,octo2024,openvla2024,gr00tn12025}. Yet these capabilities remain bounded by the distributions represented in the training data~\cite{rt12023,openvla2024,rax2026}. Extending these policies to new embodiments, tasks, or conditions beyond this coverage typically requires additional data collection, training, or fine-tuning~\cite{droid2024,octo2024,openvla2024,geminirobotics2025,pistar062026}. As a result, capability growth remains closely coupled to pre-collected demonstrations and the imitation distribution they define~\cite{rtx2024,rax2026}.

General-purpose multimodal models offer another route to robot adaptation through reasoning during deployment~\cite{saycan2022,codeaspolicies2023,voxposer2023,rekep2024}. By combining current visual observations with task instructions and execution feedback, these models can guide the robot’s actions and adjust its behavior as execution unfolds~\cite{innermonologue2022,voxposer2023,roboicl2026}. Through this interaction, the robot acquires \emph{execution knowledge}: the effects its actions produce under given conditions and the procedures that prove effective. For example, it may find that a thin coin remains on the table when pinched at its center but lifts when pinched at its rim. However, the robot retains this knowledge only within the model’s working context. Once the task ends, the knowledge is lost. On its next attempt, the robot may again pinch the coin at its center before rediscovering the rim, taking just as long as before.

Avoiding this repetition requires retaining execution knowledge across attempts. Yet simply replaying past actions is insufficient. The same action can yield different effects under different conditions. The robot therefore needs reusable procedures that determine each action from the current scene. Existing and concurrent approaches have begun to retain such procedures as skills and programs~\cite{voyager2023,lrll2024,embodiedevo2026,robofoundry2026,rax2026}. However, retaining a procedure does not ensure that it will handle conditions beyond those encountered in earlier attempts. When a procedure encounters such conditions, measured action effects provide evidence for deciding how to revise it. This evidence should include actions from failed attempts. Procedures and measured action effects therefore need to be retained together, so that a procedure can act without the model where it applies and be revised from this evidence where it does not. Existing methods that keep action-level evidence still leave each action to the model or a learned policy~\cite{lexp2026,robodoctor2026,robofoundry2026}, while programs that act without the model are frozen once developed~\cite{etm2026}.

To this end, we introduce \seek{}, a framework for agentic robot manipulation that enables self-evolution through execution knowledge while keeping the underlying model fixed. \seek{} retains this knowledge in two coupled forms. At the procedure level, a Self-Evolving Programmatic Policy (SEPP) encodes observation, action generation, verification, task progression, and recovery as executable procedures. During execution, these procedures use current observations to determine actions. At the action level, a Persistent Action-Effect Bank (PAEB) records the measured effect of each executed action with its execution context, including actions from failed attempts. Drawing on these records, the fixed model makes decisions or revises a procedure, and each revision retains references to its supporting evidence. Together, the two components form a continual self-improvement loop: execution supplies evidence for policy revision, and revised procedures carry the resulting capability into subsequent attempts. Once a retained procedure applies to the current scene, it can generate actions without another model request, so successive attempts reuse previously acquired behavior instead of repeating model deliberation.

We evaluate \seek{} on RoboDojo~\cite{robodojo2026}, a simulation benchmark of 42 dual-arm manipulation tasks. To isolate the contribution of retained knowledge, we compare the same system with and without it on scenes unseen during self-evolution; both conditions allow adaptation within an attempt. Across the 42 tasks, knowledge retained over ten attempts more than doubles task performance. Compared with GPT-6.2 Astra acting directly on 34 of these tasks, \seek{} with the same model raises the task score from 49.6 to 80.1 after five attempts and uses 77.6\% fewer model requests. We further measure the fraction of executed steps generated by retained procedures, and ablate PAEB and SEPP to separate the contributions of action-level evidence reuse and procedure-level policy evolution. Finally, on a smaller set of tasks, we replace the underlying model and run \seek{} on a real robot; in both settings, \seek{} again improves task performance.

\par\smallskip
\noindent\begin{tikzpicture}
\definecolor{seekcontribbg}{HTML}{F3ECFD}
\node[fill=seekcontribbg, text=black, draw=none, rounded corners=2pt,
      inner xsep=14pt, inner ysep=7pt, outer sep=0pt,
      text width=\dimexpr\linewidth-28pt\relax, align=justify] {%
\textbf{Contributions.}
We explore self-evolution through execution knowledge and introduce \textbf{SEEK}, which comprises PAEB, recording the measured effect of each executed action, and SEPP, an executable policy revised from these records. On RoboDojo, SEEK reaches an average task score of 54.4 (+27.4) and a success rate of 48.0\% (+28.3) across all 42 tasks, compared with 27.0 and 19.7\% for the same system without retained knowledge. With GPT-6.2 Astra, SEEK reaches 80.1 (+30.5) and 75.0\% (+34.3) on 34 of these tasks after five attempts, compared with 49.6 and 40.7\% for the same model acting directly.
};
\end{tikzpicture}
\par\smallskip

%% file: sections/02_related.tex
\section{Related Work}

\paragraph{Foundation Models for Robot Control.}

Foundation models support robot control through learned action policies~\cite{rt22023,openvla2024} and agents that reason during execution~\cite{saycan2022,innermonologue2022}.
General-purpose multimodal models can also act as the policy directly, deciding each action from observations and execution feedback without task-specific demonstrations or training~\cite{roboprobe2026}. The execution knowledge such an agent gains is held only in the model's working context and is lost once the task ends.

\paragraph{Reusable Programs for Robot Manipulation.}
Executable programs preserve control logic as reusable procedures. Code as Policies and ProgPrompt express robot behavior as programs that compose available perception and action interfaces~\cite{codeaspolicies2023,progprompt2022}.
PyRUA-Lean extends these interfaces with Python cells that perform local checks, retries, and selective observation without additional model requests~\cite{pyrua2026}.
These cells build on a primitive library that its authors wrote for the benchmark's dual-arm station and on operating notes, including how strictly the benchmark judges a final scene.
Its programs are written anew in each attempt, which therefore starts from what the authors supplied rather than from what earlier attempts revealed.

A retained program that acts without the model removes repeated deliberation for behavior already resolved, and one that also remains revisable during execution can be corrected as soon as it meets a condition that earlier attempts did not cover.
COAP retains its programs. Coding agents develop a shared library for perception, state tracking, and control through offline refinement from execution traces, then freeze the code for execution without the model~\cite{etm2026}.
This refinement uses state information that a robot acting from its own observations does not have. The agents compare the measured object state against simulator ground truth, and the perception code takes the object shapes published with the benchmark as constants.
Once frozen, the code can no longer be revised when it meets a condition that development did not cover, such as an object below a fixed size threshold, and each new task requires further offline development.
Voyager, PhysicalRSI, and RoboFoundry keep their skills revisable from execution~\cite{voyager2023,physicalrsi2026,robofoundry2026}, but the model remains in the loop; in RoboFoundry it is called at every step.

\begin{table}[t]
\centering
\definecolor{rwadapt}{HTML}{F3E8C8}
\definecolor{rwmem}{HTML}{F5DFD0}
\definecolor{rwauto}{HTML}{DDE7F0}
\definecolor{rwours}{HTML}{F3ECFD}
\caption{\textbf{Capabilities for cumulative self-improvement.} {\setlength{\fboxsep}{1pt}Header colors mark \colorbox{rwadapt}{adaptivity} within an attempt, \colorbox{rwmem}{memory} across attempts, and \colorbox{rwauto}{autonomy}.} \emph{Online revision} adapts ongoing execution; \emph{self-acquired knowledge} marks methods whose knowledge comes from their own attempts rather than from demonstrations, ground-truth object states from the simulator, or knowledge written by their authors for the benchmark; \emph{effect evidence} preserves action--outcome records; \emph{self-executing procedures} are retained procedures that generate actions without a model request where they apply; and \emph{executable reuse} retains acquired procedures.}
\label{tab:execution_knowledge}
\footnotesize
\renewcommand{\arraystretch}{1.12}
\let\rwplainxmark\xmark
\renewcommand{\xmark}{\textcolor{black!40}{\rwplainxmark}}

\def\rwtable{%
\begin{tabular}{lccccc}
\toprule
\textbf{Method} & \multicolumn{1}{>{\columncolor{rwadapt}[.5\tabcolsep][.5\tabcolsep]}c}{Online revision} & \multicolumn{1}{>{\columncolor{rwmem}[.5\tabcolsep][\tabcolsep]}c}{Self-acquired knowledge} & \multicolumn{1}{>{\columncolor{rwmem}[\tabcolsep][.5\tabcolsep]}c}{Effect evidence} & \multicolumn{1}{>{\columncolor{rwauto}[.5\tabcolsep][\tabcolsep]}c}{Self-executing procedures} & \multicolumn{1}{>{\columncolor{rwauto}[\tabcolsep][.5\tabcolsep]}c}{Executable reuse} \\
\midrule
RoboICL~\cite{roboicl2026} & \cmark & \xmark & \xmark & \xmark & \xmark \\
PyRUA-Lean~\cite{pyrua2026} & \cmark & \xmark & \xmark & \xmark & \xmark \\
LEXP~\cite{lexp2026} & \cmark & \cmark & \cmark & \xmark & \xmark \\
RoboDoctor~\cite{robodoctor2026} & \cmark & \cmark & \cmark & \xmark & \xmark \\
EmbodiedEvo~\cite{embodiedevo2026} & \cmark & \cmark & \xmark & \xmark & \cmark \\
HexaShell~\cite{hexashell2026} & \cmark & \cmark & \xmark & \xmark & \cmark \\
RoboFoundry~\cite{robofoundry2026} & \cmark & \cmark & \cmark & \xmark & \cmark \\
COAP~\cite{etm2026} & \xmark & \xmark & \cmark & \cmark & \cmark \\
\midrule
\rowcolor{rwours}
\textbf{\seek{} (ours)} & \cmark & \cmark & \cmark & \cmark & \cmark \\
\bottomrule
\end{tabular}}%
\setlength{\tabcolsep}{0pt}%
\sbox0{\rwtable}%
\setlength{\tabcolsep}{\dimexpr(\linewidth-\wd0)/12\relax}%
\noindent\rwtable
\end{table}

\paragraph{Retained Experience for Robot Manipulation.}
Interaction history, lessons, and execution evidence preserve what prior actions have revealed. Reflexion and ExpeL retain reflections and extracted insights, while DROC distills language corrections into reusable knowledge~\cite{reflexion2023,expel2024,droc2024}.
RoboICL conditions the model on a task demonstration and keeps observations, actions, and execution receipts only within an attempt~\cite{roboicl2026}.
LEXP and RoboDoctor carry lessons and execution records across attempts together with their supporting evidence~\cite{lexp2026,robodoctor2026}, but as context for the model. Neither retains an executable procedure, so each action is still decided by the model or by the learned policy it supervises.
EmbodiedEvo brings procedural and diagnostic knowledge together by distilling physical feedback into reusable code Skills and Lessons during exploration in a separate scene~\cite{embodiedevo2026}.
Only these distilled artifacts reach the test scene, where they are read-only. The interaction records behind them are not carried over, and the coding agent still writes each program that calls the Skills.
HexaShell develops its tools on the task itself. After each development attempt, an optimizer agent reads the attempt's log and adds or repairs task-specific tools, which are then frozen~\cite{hexashell2026}.
As with EmbodiedEvo, what carries forward is the distilled tools and notes rather than the records, and the acting agent still issues every command that calls a tool.

\seek{} couples persistent action-effect evidence with revisable executable procedures, both acquired from the robot's own attempts, allowing observed effects to directly support procedure revision and subsequent execution.
Where a retained procedure applies, it generates actions without a model request; where it meets a condition it does not cover, the fixed model revises it within the attempt.
Table~\ref{tab:execution_knowledge} compares these capabilities.

%% file: sections/04_method.tex
\section{Method}
\label{sec:method}
\begin{figure*}[t]
\centering
\includegraphics[width=\textwidth]{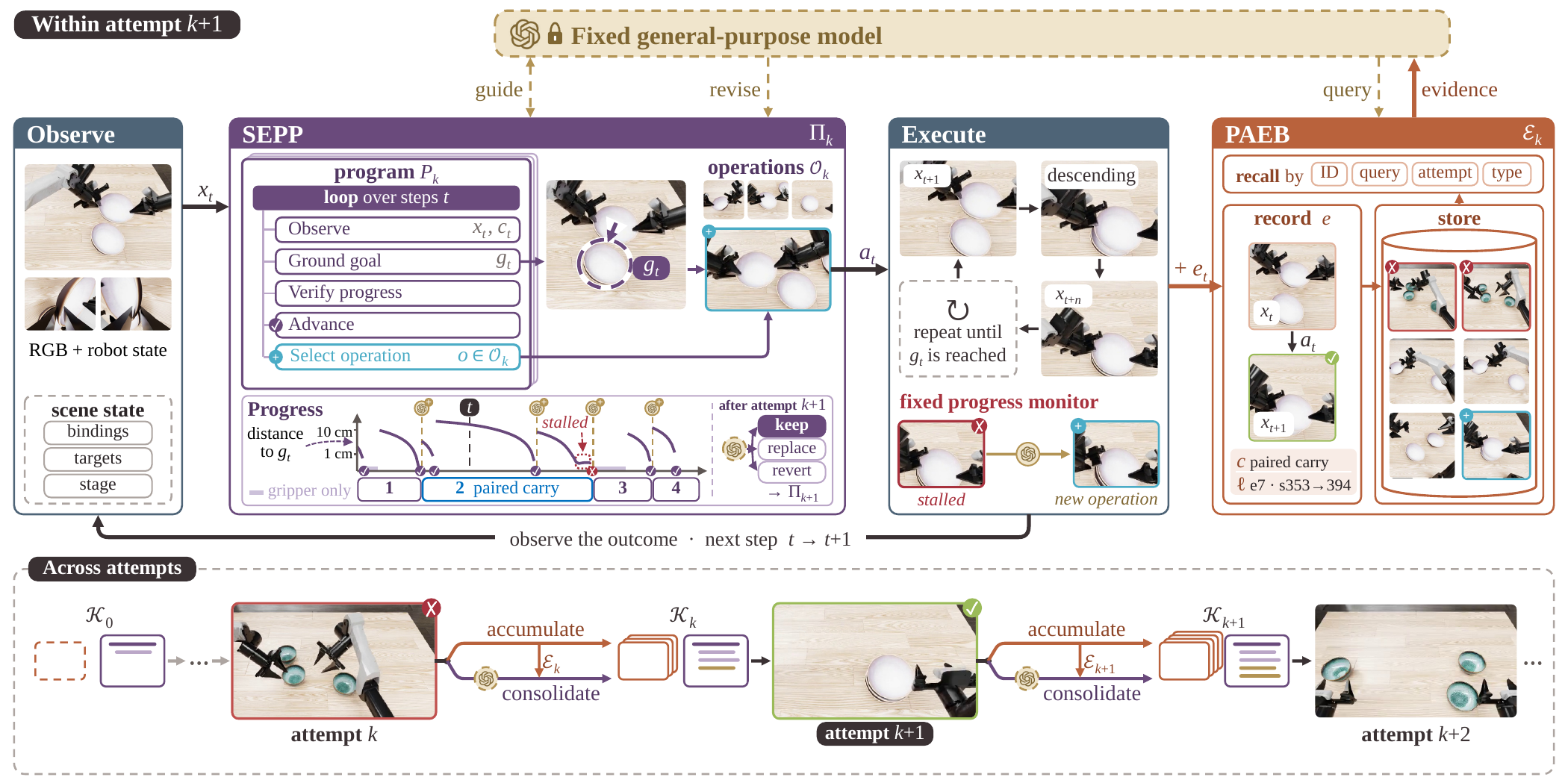}
\caption{\textbf{Execution and self-evolution in \seek.} SEPP grounds reusable procedures in the current observation $x_t$ and controls execution, while PAEB accumulates the resulting evidence. At step $t$, $c_t$ denotes the execution context used by the active procedure. In the progress panel, a check marks a goal $g_t$ that the procedure verifies by itself before advancing, and the model icons mark the only steps where the fixed model is called: when a procedure requests help or execution stalls (red cross). When called, the fixed model retrieves evidence from PAEB to guide execution, by interpreting observations or proposing action candidates, or to revise the main program or operations. Across attempts, post-attempt review consolidates the policy, while scene-dependent state is initialized anew.}
\label{fig:method_overview}
\end{figure*}

\subsection{Problem Setting}
\label{sec:formulation}
We consider repeated attempts at a manipulation task specified by a language instruction, where an attempt is one execution of the task from an initial scene layout until success or the step limit. In attempt $k$, the robot acts in a closed loop using RGB observations, robot state, and execution feedback, and may revise its behavior online. The execution trace $\tau_k$ records observations, actions, and feedback; the task-level outcome $y_k$ comprises the benchmark-reported success indicator and task score, supplied after the attempt ends. Beyond the instruction, this outcome is the only information from outside the robot's own observations, and no privileged object state is read during or after the attempt. The underlying general-purpose model remains fixed.

Across attempts, $\mathcal{K}_k$ denotes the execution knowledge retained after $k$ attempts, with $k\geq0$, and is updated by
\begin{equation}
\mathcal{K}_k=U(\mathcal{K}_{k-1},\tau_k,y_k),\qquad k\geq1,
\label{eq:knowledge_update}
\end{equation}
where $U$ incorporates the attempt's trace and outcome into the retained knowledge. Attempts that update this knowledge are evolution attempts. The initial knowledge $\mathcal{K}_0$ contains no task-specific evidence or procedures, and execution initially relies on the model's action candidates.

\subsection{Overview}
\label{sec:method_overview}
As shown in Fig.~\ref{fig:method_overview}, our proposed \seek{} retains execution knowledge in two coupled forms:
\begin{equation}
\mathcal{K}_k=\{\mathcal{E}_k,\Pi_k\},
\label{eq:persistent_state}
\end{equation}
where $\mathcal{E}_k$ contains the measured effects of executed actions and their execution context in PAEB, while $\Pi_k$ encodes executable procedures in SEPP.

During an attempt, these procedures use the current observations to determine actions and check progress. Then, each executed action adds evidence to PAEB, regardless of whether it produces the expected effect. When a procedure requests help or execution stalls, the model checks the current observations and retrieved evidence to guide the execution or revision of the procedure. After the attempt, the model reviews the trace and outcome to select the procedures to retain. The next attempt reuses these procedures and accumulated evidence, while initializing scene-dependent state from its own observations (Sec.~\ref{sec:cross_attempt}).

\subsection{Self-Evolving Programmatic Policy}
\label{sec:sepp}
In order to reuse a procedure across scene layouts, its actions must be determined based on current observations rather than fixed by previous executions. To this end, we design the Self-Evolving Programmatic Policy (SEPP), which organizes these procedures into a main program that controls task progression and reusable operations that execute behaviors:
\begin{equation}
\Pi_k=(P_k,\mathcal{O}_k),
\label{eq:sepp}
\end{equation}
where $P_k$ is the main program and $\mathcal{O}_k$ contains named operations. The main program and each operation are procedures, each defining a decision routine and optionally an observation routine, which updates the local state and evaluates the effects. The main program calls operations using scene-dependent parameters, and an operation can select another operation or return control to the main program.

To execute in the current scene, these procedures derive operation parameters and local targets from the current observations. They access RGB images, camera calibration, and robot proprioception through a shared executable interface. The fixed model binds a task-relevant object or part as a named role by giving a camera, a box in the current image, and optionally a text concept. SAM3~\cite{sam32026} segments the instance in the box, and its temporal memory propagates the mask to later images of that camera without further model requests. When the track is lost, the role is located again by matching its reference crop, confirmed by forward--backward optical flow and a new segmentation; otherwise, the role is marked invalid. Metric estimates are computed only when a procedure requests them. An image point per camera is triangulated from two or three calibrated cameras, or a single ray is intersected with a supplied plane height, and the estimate is returned with its ray residual and reprojection errors, so the procedure can test whether a correspondence is geometrically consistent. Silhouette-based estimates fit a shape hypothesis to the masks of one to three views under a supplied support height. End-effector poses and gripper measurements come from robot state. For example, the distance to $g_t$ in Fig.~\ref{fig:method_overview} compares the observed end-effector position with a specified target. An unavailable or invalid measurement is recorded as missing and never filled in from a command or an earlier step, so an objective that depends on it cannot be completed by assumption, and perception errors remain separable from action errors in the records. The interface also provides motion-feasibility queries and dispatches end-effector and gripper commands under motion constraints.

At decision step $t$, the active procedure incorporates the current observation and determines the next action $a_t$. When a numerical objective is used, $g_t$ specifies target features, tolerances, and applicability conditions. The procedure can generate an action from current measurements or select a still-valid candidate supplied by the fixed model. After dispatch, an execution receipt records the command actually executed and the robot states before and after it. The same operation and objective may span several such steps. A fixed progress monitor, separate from the revisable policy, requests model review after sustained lack of motion, oscillation, or weak response to commands.

Execution reveals which part of a procedure needs to change: target derivation, motion generation, effect assessment, operation composition, or recovery. The fixed model uses current observations and retrieved PAEB records to revise the main program, given as complete source, or selected operations, given as complete source or as literal edits against an identified stored version. Its response is checked as a whole before anything is installed: candidates against the motion limits, role bindings by starting their tracks, evidence references against the saved records, and the revised code in the interpreter, which compiles it and runs the observation and decision routines of a revised main program once on the current observation. If any check fails, nothing is installed and the error is returned with the next request; otherwise, bindings, versions, objective, and candidates are installed together, and the revised policy chooses the next action on the same observation. Procedures are written in a restricted subset of Python without imports, classes, exception handling, or file and network access, are bounded in size, run time, and call depth, and reach the robot only through the executable interface, so every motion they request passes the same executor as a model candidate. An accepted version is identified by the hash of its source and stored with its provenance; versions are therefore never overwritten, and a later review can revert a harmful one (Sec.~\ref{sec:cross_attempt}). Updating an operation preserves unrelated procedures; the current invocation keeps its version, and a subsequent invocation selects the updated one. The interpreter, measurement interfaces, motion constraints, progress monitor, and the initial routine that delegates decisions to the fixed model remain fixed runtime components; the observation and decision logic built on them can evolve.

\begin{table*}[t]
\centering
\definecolor{tabours}{HTML}{F3ECFD}
\definecolor{taboursbest}{HTML}{D6C2F5}
\definecolor{taboursecond}{HTML}{E7DAFA}
\definecolor{tabpolicy}{HTML}{DDE7F0}
\definecolor{tabagent}{HTML}{F5DFD0}
\definecolor{tabbest}{HTML}{DDDDDD}
\definecolor{tabsecond}{HTML}{EFEFEF}
\definecolor{tabsixtwo}{HTML}{F8F4FD}
\newcommand{\tbest}[1]{\cellcolor{tabbest}\textbf{#1}}
\newcommand{\tobest}[1]{\cellcolor{taboursbest}\textbf{#1}}
\newcommand{\tsec}[1]{\cellcolor{tabsecond}#1}
\newcommand{\tosec}[1]{\cellcolor{taboursecond}#1}
\newcommand{\tno}{\textcolor{black!40}{\xmark}}
\caption{\textbf{Comparison on RoboDojo.} Task score and success rate (SR, \%) by category, with category-averaged results for the full benchmark. Baselines follow the official leaderboard~\cite{robodojo2026}. The setting columns list what each method uses: \emph{Demos}, training on 100 demonstrations per task, which exist for every category except Open; and \emph{Experience}, what each method has for a task before the evaluated episodes. Shaded cells mark the best (darker, \textbf{bold}) and second-best (lighter) results; -- marks entries that do not apply or are unreported. \seek{}, no carryover: the control that starts every episode from $\mathcal{K}_0$ (Sec.~\ref{sec:cumulative_improvement}). The last block uses GPT-6.2 Astra on the 34 tasks outside Open, where the model acting directly and \seek{} after 5 attempts run the same 10 unseen episodes per task (Sec.~\ref{sec:robodojo_comparison}); its averages are over four categories and are not marked. $^{\ast}$Agent combined with a VLA. $^{\dagger}$Ten layouts, each attempted up to six times.}
\label{tab:robodojo_full}

\footnotesize
\renewcommand{\arraystretch}{1.12}
\def\maintable{%
\begin{tabular}{lcc*{12}{c}}
\toprule
& \multicolumn{2}{c}{Setting} & \multicolumn{2}{c}{Average} & \multicolumn{2}{c}{Generalization} & \multicolumn{2}{c}{Precision} & \multicolumn{2}{c}{Long-Horizon} & \multicolumn{2}{c}{Memory} & \multicolumn{2}{c}{Open} \\
\cmidrule(lr){2-3}\cmidrule(lr){4-5}\cmidrule(lr){6-7}\cmidrule(lr){8-9}\cmidrule(lr){10-11}\cmidrule(lr){12-13}\cmidrule(lr){14-15}
Method & Demos & Experience & Score & SR & Score & SR & Score & SR & Score & SR & Score & SR & Score & SR \\
\midrule
\rowcolor{tabpolicy}\multicolumn{15}{l}{\textit{VLA policies}} \\
OpenVLA-OFT~\cite{openvlaoft2025} & \cmark & -- & 0.2 & 0.0 & 0.0 & 0.0 & 0.2 & 0.0 & 0.7 & 0.0 & 0.0 & 0.0 & 0.1 & 0.1 \\
$\pi_{0}$~\cite{pi02025} & \cmark & -- & 3.5 & 1.5 & 3.9 & 2.6 & 3.6 & 0.8 & 6.2 & 2.0 & 3.5 & 2.1 & 0.3 & 0.3 \\
$\pi_{0.5}$~\cite{pi052025} & \cmark & -- & 11.4 & 6.9 & 13.4 & 8.2 & 12.4 & 5.5 & 23.5 & 14.7 & 5.9 & 4.7 & 2.0 & 1.7 \\
\addlinespace[2pt]
\rowcolor{tabpolicy}\multicolumn{15}{l}{\textit{WAM policies}} \\
InternW0-$\Delta$~\cite{internw02026} & \cmark & -- & 30.8 & 23.9 & 30.1 & 22.8 & 32.0 & 23.3 & 46.3 & 29.3 & 34.7 & 34.0 & 10.8 & 10.2 \\
VPP2~\cite{vpp22026} & \cmark & -- & 39.3 & 32.3 & 36.4 & 29.5 & \tsec{43.0} & \tsec{35.2} & \tbest{56.3} & \tsec{37.0} & 56.0 & 55.6 & 4.6 & 4.1 \\
\addlinespace[2pt]
\rowcolor{tabagent}\multicolumn{15}{l}{\textit{Agents}} \\
GPT-6 Astra~\cite{roboprobe2026} & \tno & none & 29.0 & 22.5 & 33.4 & 30.5 & 12.7 & 4.0 & 21.5 & 8.3 & 43.0 & 38.7 & 34.4 & 31.0 \\
PhysicalRSI$^{\ast}$~\cite{physicalrsi2026} & -- & -- & 36.3 & 31.4 & 21.3 & 15.6 & 38.1 & 32.5 & \tsec{46.4} & \tbest{37.3} & 46.7 & 46.6 & 28.9 & 24.9 \\
EmbodiedEvo~\cite{embodiedevo2026} & \tno & 1 scene & 45.1 & 39.0 & 35.2 & 30.8 & 25.2 & 16.3 & 33.8 & 21.8 & 79.4 & 76.7 & 51.8 & 49.3 \\

HexaShell-Preview~\cite{hexashell2026} & \tno & $\leq$60 attempts$^{\dagger}$ & \tsec{53.2} & \tsec{46.9} & \tsec{39.2} & \tsec{34.5} & 41.3 & 32.0 & 42.9 & 31.5 & \tbest{84.5} & \tbest{81.0} & \tsec{58.0} & \tsec{55.3} \\
\midrule
\seek{}, no carryover & \tno & none & 27.0 & 19.7 & 27.9 & 23.3 & 14.8 & 4.3 & 21.7 & 8.3 & 39.2 & 35.0 & 31.3 & 27.5 \\
\rowcolor{tabours}
\seek{} (ours) & \tno & 10 attempts & \tobest{54.4} & \tobest{48.0} & \tobest{42.0} & \tobest{36.3} & \tobest{43.2} & \tobest{35.3} & 44.2 & 30.8 & \tosec{79.6} & \tosec{77.0} & \tobest{63.1} & \tobest{60.8} \\
\midrule
\rowcolor{tabsixtwo}\multicolumn{15}{l}{\textit{GPT-6.2 Astra, 34 tasks outside Open}} \\
GPT-6.2 Astra~\cite{roboprobe2026} & \tno & none & 49.6 & 40.7 & 49.0 & 44.2 & 27.5 & 17.5 & 47.1 & 26.2 & 75.0 & 75.0 & -- & -- \\
\rowcolor{tabours}
\seek{} (ours) & \tno & 5 attempts & 80.1 & 75.0 & 74.8 & 69.2 & 79.2 & 75.0 & 72.8 & 62.5 & 93.6 & 93.3 & -- & -- \\
\bottomrule
\end{tabular}}%
\setlength{\tabcolsep}{0pt}%
\sbox0{\maintable}%
\setlength{\tabcolsep}{\dimexpr(\linewidth-\wd0)/30\relax}%
\noindent\maintable
\end{table*}

\subsection{Persistent Action-Effect Bank}
\label{sec:paeb}
Once its working context is lost, the fixed model no longer knows which actions have been tried or what each produced, and may repeat an ineffective action without recognizing it. PAEB retains this knowledge outside the model as a bank of the measured effects of executed actions, effective and ineffective alike, and supplies the evidence for revising procedures. For each executed action within an attempt, it records a transition
\begin{equation}
e_t=(x_t,a_t,x_{t+1},c_t,\ell_t),
\label{eq:evidence_item}
\end{equation}
where $x_t$ and $x_{t+1}$ contain observations and measured state before and after execution, $a_t$ is the dispatched action, and $c_t$ records its execution context. The provenance $\ell_t$ links the record to its attempt, procedure version, operation invocation, and execution receipt. A multi-step operation is represented by its invocation and linked transitions, so the interval illustrated in Fig.~\ref{fig:method_overview} summarizes an execution segment. Model proposals, errors, effect assessments, and reviews retain their own record types and sources. Evidence from failed attempts remains available for diagnosing ineffective actions and recovering useful partial procedures.

Given a request $q_t$ from the fixed model or a procedure, PAEB returns
\begin{equation}
\mathcal{R}_t=\operatorname{Recall}(\mathcal{E},q_t),
\label{eq:evidence_select}
\end{equation}
where $\mathcal{E}$ is the currently available bank, comprising $\mathcal{E}_{k-1}$ and the records of the current attempt. Revising a procedure requires the records behind it, which show what each action produced in the context where it was executed. A record that only resembles the current situation may stem from a different condition and mislead the revision, so $\operatorname{Recall}$ selects records by reference rather than by resemblance. Records refer to one another by IDs that encode the attempt and the step, and each version of the main program or an operation stores the IDs of the records cited when it was written, so the evidence behind a revision stays attached to it. Source-attempt and record-type filters first fix the eligible records; a request that matches nothing returns an empty result rather than records of other attempts or types. Requested IDs then select exact evidence within this scope, and a transition also carries the IDs of its observations before and after the action, so citing an observation retrieves the records made at it. A textual query, the entry point for records whose IDs are not yet known, ranks the remaining eligible records by the number of distinct tokens they share with it, with ties broken toward the most recent record. Requested IDs that are unknown, filtered out, or beyond the result limit are reported separately, so the model can tell missing evidence from evidence outside its request. The fixed model can then follow the links in the returned records, for example from an operation version to the transitions it cites, and refine its request. A request never contains the whole bank. It carries the source of the active procedure, the last 8 transitions and up to 12 earlier ones, the last three reviews, and the current and cited images, while older records appear as references that are resolved on demand, so its size stays bounded as PAEB grows.

\subsection{Self-Evolution across Attempts}
\label{sec:cross_attempt}
The update in Eq.~\eqref{eq:knowledge_update} combines evidence accumulation and policy consolidation. Evidence accumulation is
\begin{equation}
\mathcal{E}_k=\mathcal{E}_{k-1}\mathbin{\oplus}\Delta\mathcal{E}(\tau_k,y_k),
\label{eq:evidence_update}
\end{equation}
where $\oplus$ appends execution records and associated assessments and outcomes while preserving earlier evidence. Policy consolidation is
\begin{equation}
\Pi_k=\operatorname{RevisePolicy}(\Pi_{k-1},\tau_k,y_k,\mathcal{E}_k).
\label{eq:policy_update}
\end{equation}
Starting from the policy as revised online during the attempt, the fixed model reviews the outcome, executed transitions, errors, operation sequences, and available versions. It selects \emph{keep}, \emph{replace}, or \emph{revert} for the main program, and can independently update or restore selected operations. The decision is guided by recurring execution difficulties, contrasting action effects, and the observed use of existing procedures. To support this review, the fixed model can inspect historical observations and replay procedures on recorded inputs. The review does not judge a version by the final score alone. Its input groups the transitions by program version and action source, lists errors and model requests by reason, and joins each of the last three reviews with the transitions later executed under the versions it selected, so a version is assessed by what happened after it was installed, in this attempt and in later ones. A version therefore remains provisional. It is reverted, or its operation restored, when later execution provides evidence against it. The review can also write lessons, textual rules that state when they apply, what to do, what to expect, and what to do if the expectation fails; lessons are included in later requests but are not executed. A review is installed only if its code compiles, its references name saved records, and no raw record has changed; otherwise, its validation error is returned for another try, and after three failed tries the previous policy is kept.

Attempt $k+1$ inherits $\mathcal{K}_k$, while object bindings, local objectives, operation parameters, and runtime state are initialized from its own observations. A still-applicable procedure is executed without further model review, and a new challenge provides evidence for further revision.

%% file: sections/05_evidence.tex
\section{Experiments}
\label{sec:experiments}

Our experiments examine whether execution knowledge retained in PAEB and SEPP improves subsequent task performance while the underlying model remains fixed.

\subsection{Experimental Setup}
\label{sec:experimental_setup}
\paragraph{Benchmark.}
We evaluate on RoboDojo~\cite{robodojo2026}, a simulation benchmark of 42 dual-arm manipulation tasks in five capability categories: Generalization, Precision, Long-Horizon, Memory, and Open.
The robot observes RGB images from a head camera and two wrist cameras together with its proprioceptive state and current camera calibration, and acts through end-effector targets and gripper commands. Depth images and privileged scene-object states are not provided; metric visual estimates use the RGB geometry interfaces of Sec.~\ref{sec:sepp}. We report the benchmark's task score (0--100), which credits partial progress, and its strict success rate (SR), which counts only full-score episodes.

\paragraph{Model and Interface.}
The fixed model is GPT-6 Astra~\cite{gpt6astra2026} with xhigh reasoning effort.
It receives no parameter updates or task demonstrations, and the number of model requests is not capped. All tasks share the same system instructions, tools, and policy initialization.

\paragraph{Evolution and Evaluation Protocol.}
Each task starts from $\mathcal{K}_0=\{\mathcal{E}_0,\Pi_0\}$, with no task-specific evidence or procedure. Over 10 evolution attempts on distinct layouts, \seek{} updates its knowledge after each attempt by Eq.~\eqref{eq:knowledge_update} to obtain $\mathcal{K}_{10}$. The main results evaluate this snapshot over 50 episodes per task on layouts unseen during evolution, together with a no-carryover control that starts the same episodes from $\mathcal{K}_0$. Each evaluation episode is a single attempt that starts from a fresh working copy of its designated snapshot, allows feedback, evidence collection, and online revision within the episode, and discards all updates at its end. Generalization tasks use 25 standard and 25 randomized episodes. Unless noted otherwise, \seek{} refers to the system with $\mathcal{K}_{10}$.

\paragraph{Metrics.}
\emph{Operation reuse} is the fraction of executed steps generated by operations in $\mathcal{O}_k$ retained from earlier attempts. \emph{Model requests} are counted per episode, whether for scene interpretation, action candidates, or procedure revisions.

\paragraph{Analysis Tasks.}
For the analyses in Secs.~\ref{sec:cumulative_improvement}--\ref{sec:ablation}, we randomly select one or two tasks from each category, yielding nine tasks: Stack Bowls, Insert Tubes, Classify Objects, Swap T, Classify Objects by Language, Make Toast, Deposit Coin, Organize Table, and Match and Pick from Conveyor. On this subset, we also evaluate the snapshot $\mathcal{K}_5$ and two configurations that carry only evidence or only procedures across attempts, using the same episodes and protocol.

\subsection{Comparison on RoboDojo}
\label{sec:robodojo_comparison}

Table~\ref{tab:robodojo_full} compares \seek{} with learned policies and agent-based methods across all 42 tasks and lists the setting of each method. \seek{} reaches an average score of 54.4 and an SR of 48.0\%. Among the listed methods, it is highest on the average and in Generalization and Open, level with VPP2 in Precision (43.2 against 43.0), and second in Memory; in Long-Horizon, its score (44.2) is below those of VPP2 (56.3), PhysicalRSI, and InternW0-$\Delta$. \seek{} obtains these results from ten attempts per task, without demonstrations or parameter updates, and its margin over the learned policies is largest in Open (63.1 against at most 10.8), the one category for which no demonstrations exist. Table~\ref{tab:task_level_results} reports scores for 20 representative tasks, four from each category; \seek{} attains the highest score on 10 of them. These results compare the capability achieved under each method's protocol, with GPT-6 Astra at medium effort. At the xhigh effort of \seek{}, GPT-6 Astra averages 38.5 on ten tasks (five episodes each)~\cite{galbot2026astra}, against 38.7 for \seek{} without carryover and 49.9 with it.

\begin{figure}[t]
\centering
\definecolor{appgold}{HTML}{B39455}
\definecolor{appgoldfill}{HTML}{F0E5CC}
\definecolor{appgoldtext}{HTML}{7F6632}
\definecolor{appplum}{HTML}{6A4A7C}
\definecolor{appfaint}{HTML}{ADA5A1}
\definecolor{appgrid}{HTML}{DDD6D1}
\pgfplotstableread{
x sa ss ra rs
0 83.0 100.0 66.3 4.5
1 100.0 100.0 40.2 1.8
2 29.5 75.0 87.5 34.2
3 92.0 90.0 41.0 2.6
4 26.5 49.0 74.8 47.7
5 50.0 60.0 80.8 41.4
6 10.0 60.0 24.9 10.7
7 20.0 60.0 122.2 38.1
8 75.0 87.0 70.4 8.6
9 30.0 100.0 85.4 12.6
10 6.0 16.0 58.0 51.5
11 66.0 100.0 39.6 9.5
13 60.0 95.0 88.4 4.2
14 20.0 90.0 38.9 14.3
15 12.0 80.0 35.5 12.6
16 10.0 90.0 36.6 15.4
17 20.0 80.0 67.1 8.9
18 24.0 84.0 27.0 3.6
19 15.0 15.0 27.1 21.8
20 59.0 100.0 92.3 8.3
22 10.0 47.5 69.7 37.9
23 94.0 100.0 63.1 12.5
24 64.5 92.5 42.5 20.3
25 90.0 100.0 120.7 0.0
26 13.5 57.5 69.4 12.5
27 62.5 92.5 95.2 27.1
28 10.0 0.0 90.6 13.1
29 32.0 92.0 143.5 21.6
31 70.0 100.0 68.4 0.1
32 100.0 90.0 36.2 9.9
33 90.0 100.0 44.9 3.0
34 60.0 100.0 37.3 0.7
35 40.0 90.0 36.4 3.6
36 90.0 81.5 66.4 14.5
}\appdirecttable
\begin{tikzpicture}
\pgfplotsset{
  appaxis/.style={scale only axis, width=\dimexpr\linewidth-36pt\relax, height=0.95in, xmin=-0.9, xmax=42.4,
    axis y line*=left, y axis line style={draw=appfaint, line width=0.4pt}, ytick style={draw=appfaint, line width=0.4pt},
    major tick length=2pt, ymajorgrids, major grid style={draw=appgrid, line width=0.3pt},
    tick label style={font=\scriptsize}, ylabel style={font=\scriptsize}, clip=false},
  appdirect/.style={only marks, mark=*, mark size=1.7pt, color=appgold, mark options={draw=appgold, fill=appgoldfill, line width=0.5pt}},
  appseek/.style={only marks, mark=*, mark size=1.55pt, color=appplum, mark options={draw=white, fill=appplum, line width=0.2pt}},
}
\begin{axis}[appaxis, name=score, axis x line=none, ymin=-6, ymax=106, ytick={0,50,100}, ylabel={Task score $\uparrow$},
  legend style={at={(0,1)}, anchor=south west, yshift=17pt, draw=none, fill=none, legend columns=2, font=\scriptsize, inner sep=0pt,
    /tikz/every even column/.append style={column sep=9pt}}, legend cell align=left]
\addplot[forget plot, draw=appplum!55, line width=0.8pt, quiver={u=0, v=\thisrow{ss}-\thisrow{sa}}] table[x=x, y=sa]{\appdirecttable};
\addplot[appdirect] table[x=x, y=sa]{\appdirecttable};
\addlegendentry{GPT-6.2 Astra acting directly}
\addplot[appseek] table[x=x, y=ss]{\appdirecttable};
\addlegendentry{\seek{} (GPT-6.2 Astra)}
\draw[draw=appfaint, line width=0.4pt] (axis cs:-0.35,111) -- (axis cs:11.35,111);
\node[anchor=base, inner sep=0pt, font=\scriptsize\itshape] at ([yshift=3.4pt]axis cs:5.5,111) {Generalization};
\draw[draw=appfaint, line width=0.4pt] (axis cs:12.65,111) -- (axis cs:20.35,111);
\node[anchor=base, inner sep=0pt, font=\scriptsize\itshape] at ([yshift=3.4pt]axis cs:16.5,111) {Precision};
\draw[draw=appfaint, line width=0.4pt] (axis cs:21.65,111) -- (axis cs:29.35,111);
\node[anchor=base, inner sep=0pt, font=\scriptsize\itshape] at ([yshift=3.4pt]axis cs:25.5,111) {Long-Horizon};
\draw[draw=appfaint, line width=0.4pt] (axis cs:30.65,111) -- (axis cs:36.35,111);
\node[anchor=base, inner sep=0pt, font=\scriptsize\itshape] at ([yshift=3.4pt]axis cs:33.5,111) {Memory};
\draw[-{Stealth[length=4.2pt, width=4.2pt]}, draw=appplum, line width=1.1pt, shorten <=3.2pt] (axis cs:38.6,49.6) -- (axis cs:38.6,80.1);
\draw[draw=appgold, fill=appgoldfill, line width=0.8pt] (axis cs:38.6,49.6) circle[radius=2.3pt];
\node[anchor=west, inner sep=0pt, font=\scriptsize, text=appgoldtext] at ([xshift=5.5pt]axis cs:38.6,49.6) {49.6};
\node[anchor=west, inner sep=0pt, font=\scriptsize\bfseries, text=appplum] at ([xshift=5.5pt]axis cs:38.6,80.1) {80.1};
\end{axis}
\begin{axis}[appaxis, at={(score.south west)}, anchor=north west, yshift=-13pt, ymin=-9, ymax=159, ytick={0,50,100,150},
  ylabel={Requests per episode $\downarrow$}, axis x line*=bottom, x axis line style={draw=none}, xtick style={draw=none},
  xtick={0,1,2,3,4,5,6,7,8,9,10,11,13,14,15,16,17,18,19,20,22,23,24,25,26,27,28,29,31,32,33,34,35,36,38.6},
  xticklabels={Stack Bowls,Push T,Pack Objects into Box,Fold Clothes,Hang Mugs,Sweep Blocks,Pour Liquid into Cup,Make Toast,Arrange Largest Number,Sort Nesting Dolls by Size,Store Laptop and Headphones,Stack Blocks,Fasten Screws,Plug in Charger,Insert Tubes,Pour Balls into Vase,Play Xylophone,Deposit Coin,Insert Key,Build Tower,Fill Pen Holder,Classify Objects,Put Bottles into Dustbin,Play Tic-Tac-Toe,Fill Egg Holder,Organize Table,Make Kong,Play Stacking Toy,Cover Blocks,Match and Pick from Conveyor,Swap Blocks,Swap T,Press by Number,Imitate Sorting Sequence,\textbf{Average}},
  x tick label style={rotate=50, anchor=east, font=\fontsize{6}{7}\selectfont, inner sep=1.5pt}]
\addplot[forget plot, draw=appplum!55, line width=0.8pt, quiver={u=0, v=\thisrow{rs}-\thisrow{ra}}] table[x=x, y=ra]{\appdirecttable};
\addplot[appdirect] table[x=x, y=ra]{\appdirecttable};
\addplot[appseek] table[x=x, y=rs]{\appdirecttable};
\draw[-{Stealth[length=4.2pt, width=4.2pt]}, draw=appplum, line width=1.1pt, shorten <=3.2pt] (axis cs:38.6,63.2) -- (axis cs:38.6,14.1);
\draw[draw=appgold, fill=appgoldfill, line width=0.8pt] (axis cs:38.6,63.2) circle[radius=2.3pt];
\node[anchor=west, inner sep=0pt, font=\scriptsize, text=appgoldtext] at ([xshift=5.5pt]axis cs:38.6,63.2) {63.2};
\node[anchor=west, inner sep=0pt, font=\scriptsize\bfseries, text=appplum] at ([xshift=5.5pt]axis cs:38.6,14.1) {14.1};
\end{axis}
\end{tikzpicture}
\caption{\textbf{Per-task change from GPT-6.2 Astra acting directly to \seek{}.} On the 34 tasks outside Open, a line joins the direct agent and \seek{} after 5 evolution attempts, both with GPT-6.2 Astra, in task score (top) and in model requests per episode (bottom). The last column gives the averages.}
\label{fig:direct62}
\end{figure}
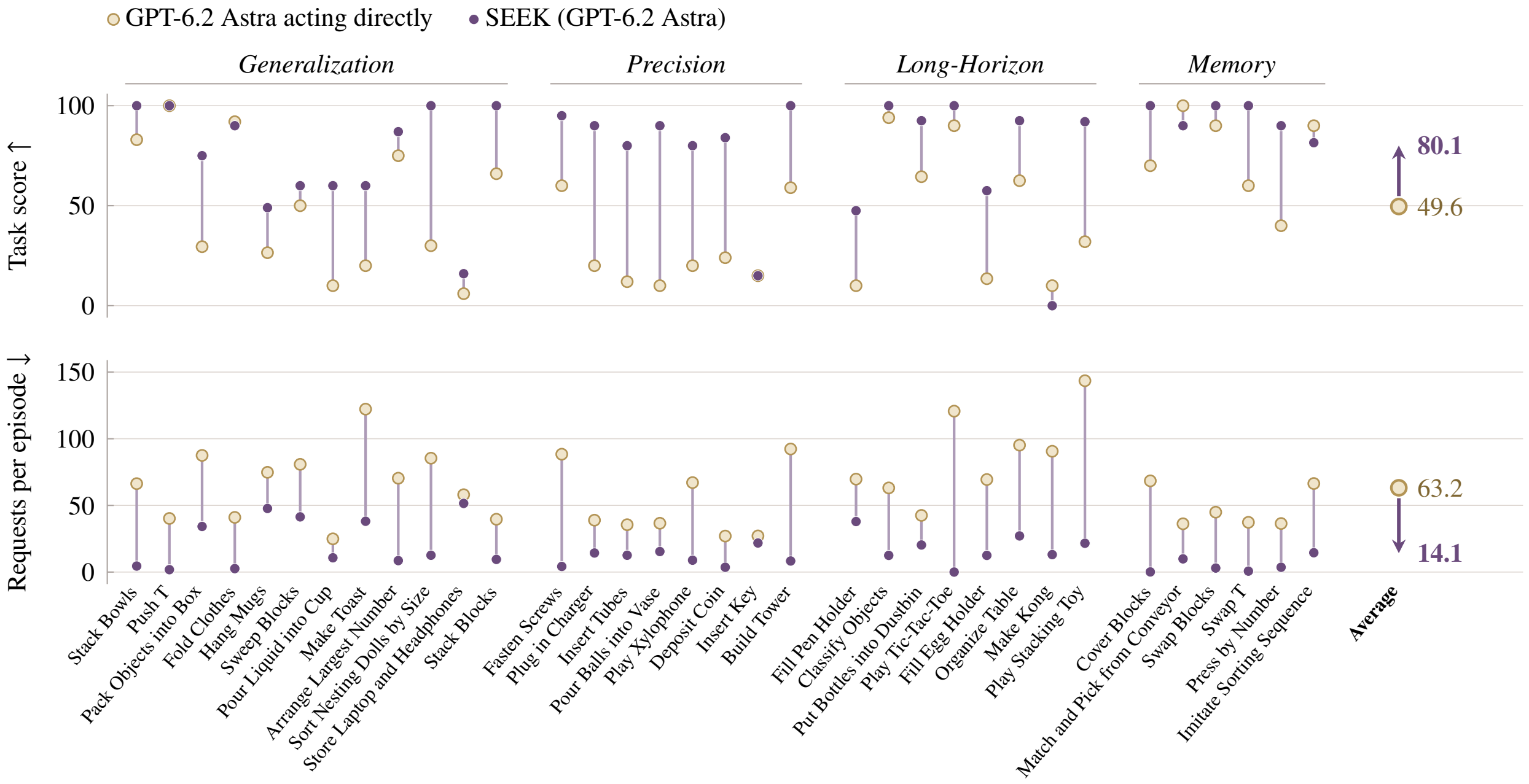

\paragraph{GPT-6.2 Astra.}
We also evaluate \seek{} with GPT-6.2 Astra at xhigh effort on the 34 tasks outside Open. For each task, \seek{} runs 5 evolution attempts on distinct layouts, and the resulting $\mathcal{K}_5$ is evaluated over 10 episodes on unseen layouts under the protocol of Sec.~\ref{sec:experimental_setup}, with 5 standard and 5 randomized episodes for Generalization tasks. GPT-6.2 Astra acting directly runs the same episodes through the reference harness of RoboProbe~\cite{roboprobe2026}. \seek{} reaches an average score of 80.1 and an SR of 75.0\%, against 49.6 and 40.7\% for the direct agent (Table~\ref{tab:robodojo_full} and Fig.~\ref{fig:direct62}), and uses 14.1 model requests per episode, against 63.2 (77.6\% fewer).

\begin{table}[t]
\centering
\definecolor{tabcat}{HTML}{F3ECFD}
\definecolor{tabbest}{HTML}{DDDDDD}
\definecolor{tabsecond}{HTML}{EFEFEF}
\definecolor{tabsixtwo}{HTML}{F8F4FD}
\newcommand{\tbest}[1]{\cellcolor{tabbest}\textbf{#1}}
\newcommand{\tsec}[1]{\cellcolor{tabsecond}#1}
\caption{\textbf{Task-level scores on 20 representative RoboDojo tasks.} Settings, shading, and symbols follow Table~\ref{tab:robodojo_full}; the best and second-best results are marked per task. Columns shaded in lighter purple use GPT-6.2 Astra, as in the last block of Table~\ref{tab:robodojo_full}, and are unmarked.}
\label{tab:task_level_results}
\scriptsize
\renewcommand{\arraystretch}{1.12}
\def\tasktable{%
\begin{tabular}{l!{\hspace{1em}}*{5}{c}*{2}{>{\columncolor{tabsixtwo}}c}}
\toprule
Task & GPT-6 Astra & EmbodiedEvo & PhysicalRSI$^{\ast}$ & HexaShell-Preview & \seek{} (ours) & GPT-6.2 Astra & \seek{} \\
\midrule
\rowcolor{tabcat}\multicolumn{8}{l}{\textit{Generalization}} \\
Make Toast & 0.5 & 5.0 & \tsec{5.3} & 1.5 & \tbest{25.0} & 20.0 & 60.0 \\
Stack Bowls & 66.5 & 59.9 & 44.2 & \tbest{86.9} & \tsec{74.9} & 83.0 & 100.0 \\
Push T & 60.0 & 86.0 & 32.7 & \tbest{96.0} & \tsec{88.0} & 100.0 & 100.0 \\
Pack Objects into Box & 11.6 & 14.1 & 20.7 & \tbest{33.1} & \tsec{21.7} & 29.5 & 75.0 \\
\addlinespace[2pt]
\rowcolor{tabcat}\multicolumn{8}{l}{\textit{Precision}} \\
Build Tower & 16.4 & 9.0 & 37.1 & \tbest{48.0} & \tsec{38.2} & 59.0 & 100.0 \\
Deposit Coin & 16.0 & 37.6 & \tbest{62.3} & 26.4 & \tsec{40.8} & 24.0 & 84.0 \\
Insert Key & 14.4 & 15.0 & \tbest{48.4} & 15.0 & \tsec{20.1} & 15.0 & 15.0 \\
Insert Tubes & 6.0 & 46.0 & \tsec{78.9} & 40.8 & \tbest{80.0} & 12.0 & 80.0 \\
\addlinespace[2pt]
\rowcolor{tabcat}\multicolumn{8}{l}{\textit{Long-Horizon}} \\
Classify Objects & 69.0 & 35.5 & 55.8 & \tsec{85.3} & \tbest{86.5} & 94.0 & 100.0 \\
Fill Egg Holder & 5.5 & \tsec{15.8} & 1.9 & 8.8 & \tbest{33.9} & 13.5 & 57.5 \\
Organize Table & \tsec{39.5} & 35.0 & 27.3 & 14.0 & \tbest{42.5} & 62.5 & 92.5 \\
Fill Pen Holder & 5.0 & 0.9 & \tbest{21.9} & 6.1 & \tsec{8.0} & 10.0 & 47.5 \\
\addlinespace[2pt]
\rowcolor{tabcat}\multicolumn{8}{l}{\textit{Memory}} \\
Imitate Sorting Sequence & 58.9 & \tbest{75.8} & 1.1 & \tsec{63.2} & 59.9 & 90.0 & 81.5 \\
Match and Pick from Conveyor & 62.0 & 60.0 & 7.3 & \tsec{74.0} & \tbest{94.0} & 100.0 & 90.0 \\
Press by Number & 70.0 & 88.0 & 79.3 & \tsec{96.0} & \tbest{100.0} & 40.0 & 90.0 \\
Swap T & 18.0 & 88.0 & 92.7 & \tbest{96.0} & \tsec{94.0} & 60.0 & 100.0 \\
\addlinespace[2pt]
\rowcolor{tabcat}\multicolumn{8}{l}{\textit{Open}} \\
Align Blocks & 50.0 & \tbest{98.0} & 86.0 & 78.0 & \tsec{92.0} & -- & -- \\
Classify Objects by Language & 46.0 & 58.0 & 20.9 & \tsec{79.4} & \tbest{95.6} & -- & -- \\
General Pickup & \tsec{84.0} & \tsec{84.0} & 47.3 & 82.0 & \tbest{96.0} & -- & -- \\
Pick from Conveyor by Image & 4.0 & \tsec{52.0} & 0.0 & 28.0 & \tbest{56.0} & -- & -- \\
\bottomrule
\end{tabular}}%
\setlength{\tabcolsep}{0pt}%
\sbox0{\tasktable}%
\setlength{\tabcolsep}{\dimexpr(\linewidth-\wd0)/16\relax}%
\noindent\tasktable
\end{table}

\subsection{Cumulative Self-Improvement}
\label{sec:cumulative_improvement}
We compare \seek{} with the no-carryover control of Sec.~\ref{sec:experimental_setup}. Both run the same unseen episodes with the same within-episode adaptation, so their difference measures the contribution of retained knowledge.

Across all 42 tasks, retained knowledge raises the average score from 27.0 to 54.4 and the SR from 19.7\% to 48.0\% (Table~\ref{tab:robodojo_full}), with gains in every category and on 37 of the 42 tasks. On the nine analysis tasks (Table~\ref{tab:ablation}, first and last configurations), it raises the average score from 30.3 to 70.4, with gains on all nine tasks and the largest on Swap T (18.0 to 94.0). Model requests decrease from 54.4 to 41.5 per episode (23.7\%). The snapshot after five attempts, $\mathcal{K}_5$, reaches 53.9 with 51.4 requests per episode. The first five attempts thus bring 23.6 of the 40.1 points, while most of the reduction in model requests comes in the next five.

\seek{} also exceeds GPT-6 Astra acting directly at medium effort~\cite{roboprobe2026}, which averages 35.9 on the nine analysis tasks, and is higher on all nine. The control itself scores below this agent (30.3). Without retained knowledge, it must write its procedures within the episode, and we observe that these first versions take more steps than actions issued directly by the model and often exhaust the task-specific step limit.

\begin{table}[t]
\centering
\definecolor{tabours}{HTML}{F3ECFD}
\newcommand{\tno}{\textcolor{black!40}{\xmark}}
\newcommand{\ablcfg}[1]{\multicolumn{2}{c}{#1}}
\caption{\textbf{Effect of retained knowledge and of its two components.} Req.\ (shaded): model requests per episode. The last two configurations carry both, after 5 and 10 evolution attempts.}

\label{tab:ablation}
\footnotesize
\renewcommand{\arraystretch}{1.12}
\def\abltable{%
\begin{tabular}{l!{\hspace{2em}}*{5}{c>{\columncolor{tabours}}c}}
\toprule
PAEB carryover & \ablcfg{\tno} & \ablcfg{\tno} & \ablcfg{\cmark} & \ablcfg{\cmark} & \ablcfg{\cmark} \\
SEPP carryover & \ablcfg{\tno} & \ablcfg{\cmark} & \ablcfg{\tno} & \ablcfg{\cmark} & \ablcfg{\cmark} \\
Evolution attempts & \ablcfg{--} & \ablcfg{10} & \ablcfg{10} & \ablcfg{5} & \ablcfg{10} \\
\cmidrule(lr){2-3}\cmidrule(lr){4-5}\cmidrule(lr){6-7}\cmidrule(lr){8-9}\cmidrule(l){10-11}
Task & Score & Req. & Score & Req. & Score & Req. & Score & Req. & Score & Req. \\
\midrule
Stack Bowls & 33.3 & 42.0 & 57.5 & 33.9 & 64.4 & 38.6 & 56.1 & 36.5 & \textbf{74.9} & \textbf{27.9} \\
Insert Tubes & 13.2 & 58.5 & 43.2 & 45.5 & 67.6 & 43.1 & 51.2 & 45.5 & \textbf{80.0} & \textbf{33.8} \\
Classify Objects & 59.7 & 50.7 & 75.5 & 38.8 & 78.3 & 41.8 & 70.7 & 43.9 & \textbf{86.5} & \textbf{34.9} \\
Swap T & 18.0 & 30.4 & 30.0 & 32.0 & 62.0 & 18.1 & 60.0 & 22.9 & \textbf{94.0} & \textbf{15.7} \\
Classify Objects by Language & 25.2 & 54.7 & 57.2 & 47.1 & 71.8 & 46.0 & 71.0 & 52.3 & \textbf{95.6} & \textbf{38.1} \\
Make Toast & 7.0 & 100.8 & 16.5 & 91.2 & 19.5 & 94.2 & 24.5 & 101.8 & \textbf{25.0} & \textbf{89.4} \\
Deposit Coin & 16.4 & 45.0 & 29.6 & 35.9 & 33.6 & 40.0 & 24.8 & 38.5 & \textbf{40.8} & \textbf{31.9} \\
Organize Table & 36.0 & 81.3 & 38.0 & \textbf{79.1} & 40.5 & 86.7 & \textbf{43.0} & 97.7 & 42.5 & 84.5 \\
Match and Pick from Conveyor & 64.0 & 25.9 & 80.0 & 20.2 & 78.0 & 22.3 & 84.0 & 23.1 & \textbf{94.0} & \textbf{17.2} \\
\midrule
Average & 30.3 & 54.4 & 47.5 & 47.1 & 57.3 & 47.9 & 53.9 & 51.4 & \textbf{70.4} & \textbf{41.5} \\
\midrule
Operation reuse (\%) & \ablcfg{N/A} & \ablcfg{68.4} & \ablcfg{N/A} & \ablcfg{75.8} & \ablcfg{82.3} \\
\bottomrule
\end{tabular}}%
\setlength{\tabcolsep}{0pt}%
\sbox0{\abltable}%
\setlength{\tabcolsep}{\dimexpr(\linewidth-\wd0)/22\relax}%
\noindent\abltable
\end{table}

\begin{figure*}[t]
\centering
\includegraphics[width=\textwidth]{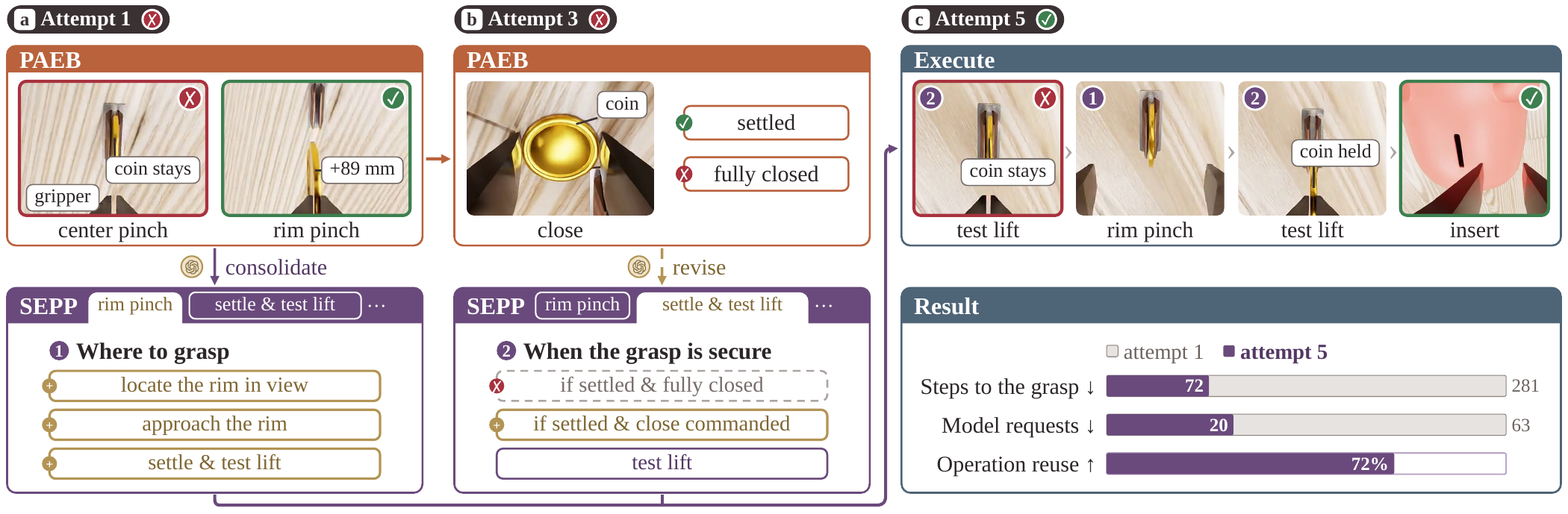}
\caption{\textbf{Case study on Deposit Coin.} (a)~After attempt~1, consolidation creates the rim-pinch operation (\textcircled{\scriptsize 1}) from PAEB evidence. (b)~Grasp verification (\textcircled{\scriptsize 2}) is revised online in attempt~3 and retained. (c)~Attempt~5 reuses both, with a test lift detecting an empty grasp (\textcircled{\scriptsize 2}) and triggering a rim re-grasp (\textcircled{\scriptsize 1}). Numbered frames mark the supporting procedure. Orange and purple arrows carry evidence and procedures to later attempts. In each SEPP card, the open tab is the changed operation, with additions in gold and removals dashed.}
\label{fig:case_study}

\end{figure*}

\subsection{Ablation Study}
\label{sec:ablation}
We separate action-level evidence reuse from procedure-level policy evolution with a $2\times2$ ablation of the two cross-attempt updates, evidence accumulation (Eq.~\eqref{eq:evidence_update}) and policy consolidation (Eq.~\eqref{eq:policy_update}). Evidence collection and online revision within an attempt remain available in every configuration; the configurations differ only in what is carried to the next attempt. Disabling both updates gives the no-carryover control of Sec.~\ref{sec:cumulative_improvement}; enabling both gives \seek{} with $\mathcal{K}_{10}$. Each of the other two configurations repeats the 10 evolution attempts with one update disabled and is evaluated from its own snapshot after the tenth attempt.

Without PAEB carryover, the bank is emptied after every attempt ($\mathcal{E}_k=\emptyset$), but post-attempt review still receives that attempt's trace $\tau_k$ and outcome $y_k$; this configuration retains procedure-level policy evolution alone. Without SEPP carryover, each attempt starts from $\Pi_0$ and post-attempt review is disabled, while PAEB accumulates evidence across attempts; this configuration retains action-level evidence reuse alone. Procedures can still be generated and revised within the current attempt, but are discarded before the next attempt.

Relative to the control (Table~\ref{tab:ablation}), procedure-level policy evolution alone raises the average score from 30.3 to 47.5, and action-level evidence reuse alone raises it to 57.3. In the first case, a procedure that meets a new condition is revised without the records of earlier attempts; in the second, the model knows what earlier actions produced but must turn this into procedures again in every attempt. With both, \seek{} reaches 70.4 and a higher score than either alone on every task, so the two components are complementary. Each alone reduces model requests from 54.4 to below 48 per episode; together they reduce them to 41.5. Operation reuse, defined only when procedures are carried, also rises with PAEB carryover, from 68.4\% to 82.3\% on average.

\subsection{Case Study}
\label{sec:case_study}

Deposit Coin is a Precision task. The robot picks up a thin coin lying flat on a table and inserts it into a piggy bank. Fig.~\ref{fig:case_study} traces the first five attempts, linking PAEB evidence to online revision and post-attempt consolidation in SEPP.

Attempt~1 fails, but its PAEB records reveal contrasting effects. A center pinch leaves the coin on the table, while a rim pinch raises it by 89\,mm. Consolidation turns this contrast into \textit{rim\_pinch}, a SEPP operation that locates the rim in the current observation, approaches with a pad offset, and checks the grasp with a settle-and-test lift (Fig.~\ref{fig:case_study}a, \textcircled{\scriptsize 1}).

In attempt~3, the fingers settle with a joint change below 0.001\,mm, yet grasp verification fails. The check requires full closure, which the coin prevents. At step~107, the fixed model revises the check online, citing PAEB records from attempt~1. Verification now requires a close command and settled fingers (Fig.~\ref{fig:case_study}b, \textcircled{\scriptsize 2}). Post-attempt review retains it.

Attempt~5 combines both procedures on a new layout (Fig.~\ref{fig:case_study}c). The first closure misses the coin; the revised check detects the empty test lift, \textit{rim\_pinch} re-grasps at the rim, and a confirmed lift allows transport and insertion. Relative to attempt~1, grasp acquisition takes 72 steps rather than 281, model requests fall from 63 to 20, and retained operations generate 72\% of the executed steps.

\subsection{Alternative Model and Real Robot}
\label{sec:model_real}

\paragraph{Underlying Model.}
We test whether self-evolution depends on the fixed model by replacing GPT-6 Astra with Claude Opus~5.5, leaving all other settings unchanged. On three of the analysis tasks, we repeat the 10 evolution attempts and the evaluation on unseen episodes. Retained knowledge raises the average score from 24.0 ($\mathcal{K}_0$) to 62.4 ($\mathcal{K}_{10}$) and reduces model requests from 35.7 to 26.0 per episode.

\begin{figure*}[t]
\centering
\includegraphics[width=\textwidth]{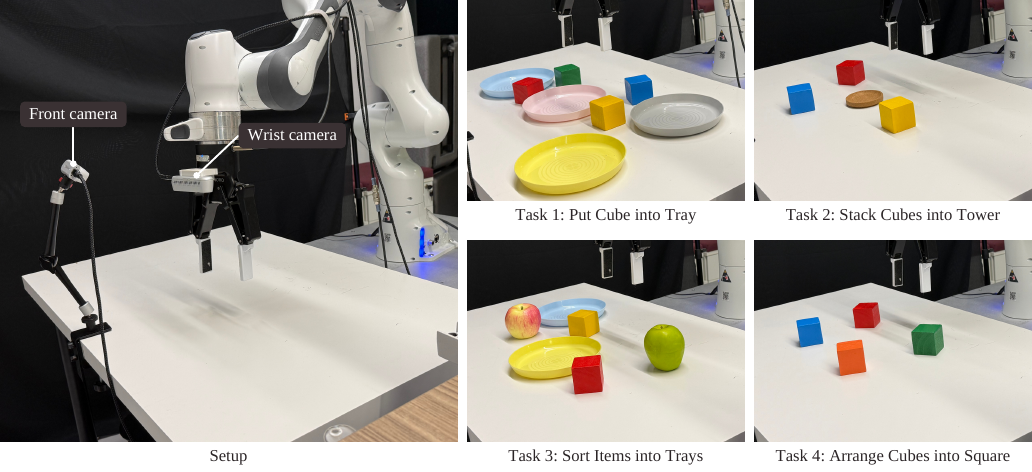}
\caption{\textbf{Real-robot setup and tasks.} The Franka arm with its wrist and front cameras, and each of the four tasks in an example layout.}
\label{fig:real_setup}
\end{figure*}

\paragraph{Real Robot.}
We test whether \seek{} improves performance under real sensing and actuation by running it on a Franka Emika Panda arm. On four tabletop tasks, including Arrange Cubes into Square (Fig.~\ref{fig:framework}), 5 evolution attempts on distinct layouts produce $\mathcal{K}_5$, which we evaluate over 10 unseen episodes per task against GPT-6 Astra acting directly at xhigh effort. Relative to this agent, \seek{} raises the average score from 66.9 to 83.3 and reduces model requests from 52.2 to 34.4 per episode.

Fig.~\ref{fig:real_setup} shows the setup and the four tasks. The arm carries a Robotiq 2F-140 gripper and uses the system instructions and tools of simulation. As in simulation, it observes RGB images, without depth images or object states, together with its proprioceptive state and camera calibration, and acts through end-effector targets and gripper commands. The images come from two Intel RealSense D435 cameras: a wrist camera looking along the fingers and a front camera fixed at the far edge of the table, facing the robot. Each episode receives a task score from 0 to 100 for the highest level of progress it reaches, and SR counts episodes with a full score. An operator scores each episode by the following rules, whose conditions are directly observable; an object is in a tray when it lies fully within the tray's rim and has been released. Task~1, \emph{Put Cube into Tray}, puts the red cube into the gray tray, with gray, pink, blue, and yellow trays and red, green, blue, and yellow cubes on the table: 100 if the red cube is in the gray tray and no other object has been placed in any tray, otherwise 0. Task~2, \emph{Stack Cubes into Tower}, stacks three cubes on a coaster in a specified order, red at the bottom, then yellow, and blue on top: 20 once the whole bottom face of the red cube rests on the coaster and the cube has been released, 50 once the yellow cube stands stably on it after release, and 100 once the blue cube stands stably on the yellow one and the whole tower stands on its own; the score is 0 if the red cube does not rest on the coaster. Task~3, \emph{Sort Items into Trays}, puts two fruits into the yellow tray and two cubes into the blue tray, in any order: 25 for each item that ends in its tray and none for an item in the other tray. Task~4, \emph{Arrange Cubes into Square}, assembles four cubes into a $2\times2$ square: 100 when all four stand directly on the table, none stacked, and form two complete rows and columns in which adjacent cubes touch, with a gap of at most about 5\,mm, and are aligned, with a rotation difference of at most about $10^\circ$ and outer faces offset by at most about 5\,mm; 60 when the square is incomplete but three cubes, touching and aligned in this way, form an L that covers three of its corners; otherwise 0.

%% file: sections/07_conclusion.tex
\section{Conclusion}
\seek{} enables self-evolution across manipulation attempts by coupling measured action-effect evidence with revisable executable procedures. PAEB retains evidence from successful and failed actions; SEPP uses this evidence to refine subsequent execution while the underlying model remains fixed. On RoboDojo, \seek{} reaches an average task score of 54.4 and a success rate of 48.0\% across all 42 tasks, where the knowledge retained over ten attempts more than doubles \seek{}'s task performance. With GPT-6.2 Astra on 34 of these tasks, \seek{} reaches 80.1 after five attempts, against 49.6 for the model acting directly, and uses 77.6\% fewer model requests.

\paragraph{Limitations.}
First, although \seek{} can revise its main program and operations, its perception tools, executor, and action interface remain fixed, so its performance still depends on these components, including the accuracy of perception models such as SAM3; stronger models could improve perception and support more precise manipulation. Second, our evaluation focuses on dual-arm manipulation in simulation and single-arm manipulation on a real Franka arm; extending \seek{} to wheeled and humanoid robots would test its generalization to more complex, long-horizon indoor and outdoor tasks. Since knowledge is accumulated separately for each task, we will also study its transfer across tasks and its accumulation over longer periods, including how to compress execution evidence, handle outdated knowledge, and maintain a growing procedure library.